\def\year{2021}\relax
\documentclass[letterpaper]{article} 
\usepackage{aaai21}  
\usepackage{times}  
\usepackage{helvet} 
\usepackage{courier}  
\usepackage[hyphens]{url}  
\usepackage{graphicx} 
\usepackage{natbib}  
\usepackage{caption} 
\usepackage{subfigure}
\usepackage{epstopdf}
\usepackage{tikz}
\usepackage{multirow}
\usepackage{footmisc}
\usepackage{amssymb}
\usepackage{multicol,multicap,graphicx}
\title{SDMTL: Semi-Decoupled Multi-grained Trajectory Learning for 3D human motion prediction}
\author {
        Xiaoli Liu,\textsuperscript{\rm 1}
        Jianqin Yin \textsuperscript{\rm 1}\thanks{Corresponding author.} \\
}
\affiliations {
    \textsuperscript{\rm 1} School of Artificial Intelligence, Beijing University of Posts and Telecommunications,
Beijing, China. \\
    Liuxiaoli134@bupt.edu.cn, jqyin@bupt.edu.cn
}

\begin{document}

\maketitle

\begin{abstract}
Predicting future human motion is critical for intelligent robots to interact with humans in the real world, and human motion has the nature of multi-granularity. However, most of the existing work either implicitly modeled multi-granularity information via fixed modes or focused on modeling a single granularity, making it hard to well capture this nature for accurate predictions. In contrast, we propose a novel end-to-end network, Semi-Decoupled Multi-grained Trajectory Learning network (SDMTL), to predict future poses, which not only flexibly captures rich multi-grained trajectory information but also aggregates multi-granularity information for predictions. Specifically, we first introduce a Brain-inspired Semi-decoupled Motion-sensitive Encoding module (BSME), effectively capturing spatiotemporal features in a semi-decoupled manner. Then, we capture the temporal dynamics of motion trajectory at multi-granularity, including fine granularity and coarse granularity. We learn multi-grained trajectory information using BSMEs hierarchically and further capture the information of temporal evolutional directions at each granularity by gathering the outputs of BSMEs at each granularity and applying temporal convolutions along the motion trajectory. Next, the captured motion dynamics can be further enhanced by aggregating the information of multi-granularity with a weighted summation scheme.
Finally, experimental results on two benchmarks, including Human3.6M and CMU-Mocap, show that our method achieves state-of-the-art performance, demonstrating the effectiveness of our proposed method. The code will be available if the paper is accepted.
\end{abstract}

\section{Introduction}
\begin{figure}[t]
\centering  
\includegraphics[width=0.8\columnwidth,height=1.3in, trim =53mm 15mm 85mm 25mm, clip=true]{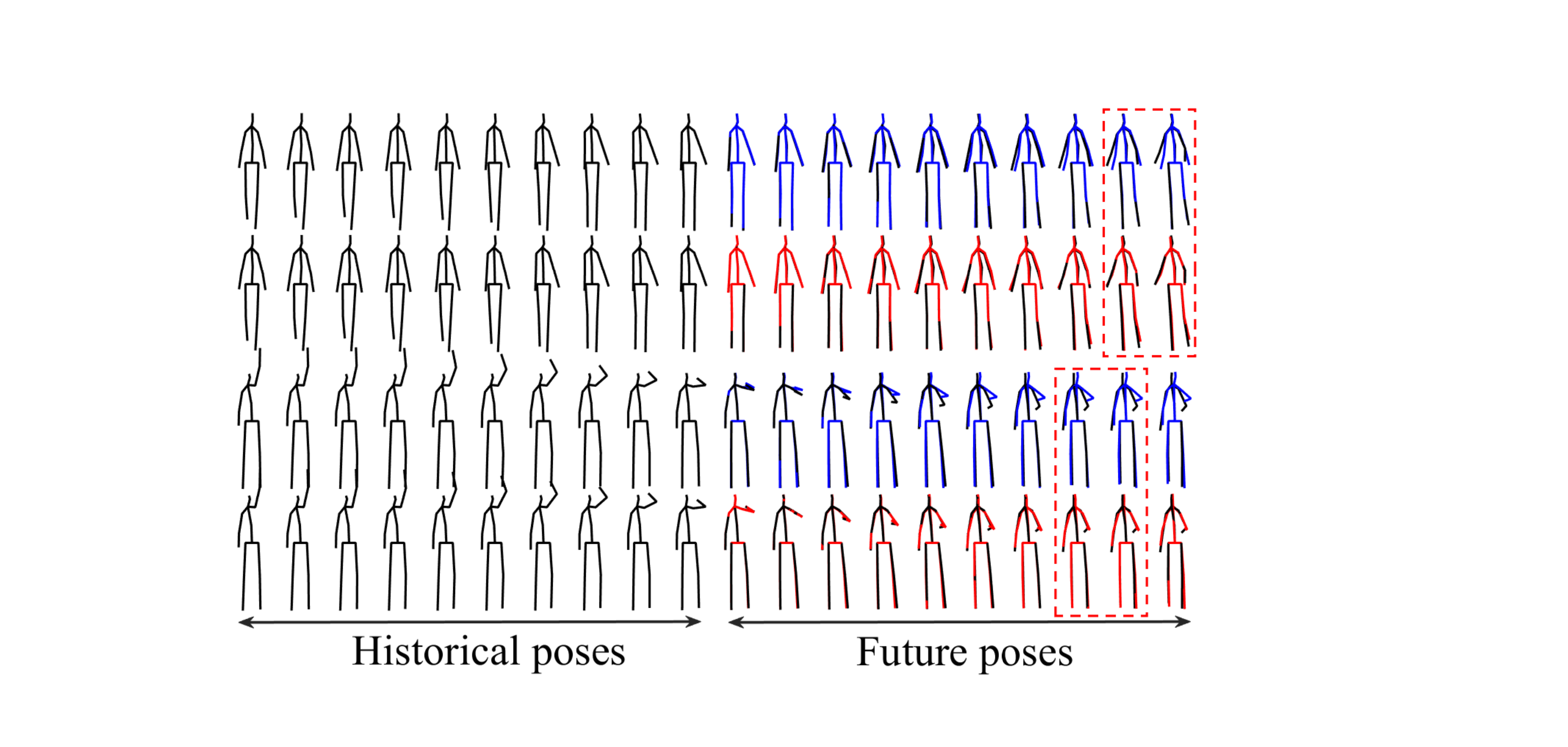} 
\caption{Human motion prediction. Here, the black poses denote the groundtruth, the blue poses denote the predictions of LTD \cite{LTD} and the red poses denote the predictions of our method (this is the same in Figure \ref{vish36m} and Figure \ref{viscmu}).}
\label{hmp}
\end{figure}

Predicting future human dynamics is of great importance in many real applications such as human-computer interaction and robotic vision \cite{bigan,ste,LTD}, enabling the robot to quickly react, prepare ahead, and interact with people. In this paper, we focus on predicting future human motion conditioning on a series of historical 3D poses, as is shown in Figure \ref{hmp}.

Traditional neural networks such as CNN (Convolutional Neural Networks) or RNN (Recurrent Neural Network) processed the spatial and temporal information in a coupled manner, and most of the exiting models for sequential tasks follow this manner \cite{hmprnn,convseq2seq,ltowards}. However, recent research shows that the human brain receives spatial and temporal representation through different pathways \cite{ecocean,scs}. This implies that the spatial and temporal features are first processed separately via different mechanisms and then are integrated together to keep the sequence of the data. Inspired by these mechanisms of the brain, we introduce a brain-inspired semi-decoupled motion-sensitive module that decouples the process of spatial and temporal modeling, i.e., first model the spatial and temporal information separately and then integrate together, for predicting human motion.

Moreover, human motion has the nature of multi-granularity and therefore needs multi-granularity modeling.
Taking a complex human activity ``hug'' as an example, it consists of multiple atomic actions such as ``approach'', ``grab'' and ``touch''. To understand each atomic action, it reflects the modeling of a fine granularity; but to understand the relationships of these atomic actions for recognizing the activity ``hug'', it reflects the modeling of coarse granularity. Moreover, different complex activities consist of various atomic actions and they reflect the modeling of various granularities. In addition, the same static pose usually occurs in different movement patterns, and thus, to distinguish these patterns, it is necessary to capture the information of temporal evolutional directions. To sum up, it is important to capture motion dynamics with rich granularities and model temporal evolutional directions for accurate predictions.

However, most of the existing methods for predicting human motion either implicitly modeled motion dynamics at multi-granularity via fixed modes, i.e., the increased temporal receptive field in CNN models \cite{convseq2seq} and the memory mechanism in RNN models \cite{rnmhy,hmprnn}, or focused on modeling motion dynamics at a single granularity such as the global time scale \cite{LTD,ste}.
The multi-grained modeling ability of these models is limit, making it hard to well capture the nature of multi-granularity and consider the information of temporal evolutional directions at each granularity over the whole sequence. For example,
\cite{LTD} focused on modeling global time scale in trajectory space via DCT (Discrete Cosine Transform) and ignored modeling motion dynamics of human motion at other granularity such as a fine granularity, failing to capture the nature of multi-granularity of human movement.
In this paper, we capture rich granularity information with adaptive scales according to the length of the input sequence and model temporal evolutional directions at each granularity, leading to more accurate predictions than fixed modes or single granularity modeling ones, as is shown in Figure \ref{hmp}.

In this paper, we address these issues by exploiting the feedforward neural network and propose a new network, SDMTL,  for capturing semi-decoupled spatiotemporal dynamics to predict future motion, modeling rich granularity information by both summarizing temporal dynamics with adaptive granularity and modeling temporal evolutional directions.
Firstly, inspired by the different cognitive mechanisms of brains for the spatial and temporal representation, we propose a brain-inspired semi-decoupled motion-sensitive module, BSME, that decouples the process of spatiotemporal modeling by first modeling the motion features that subtraction of adjacent inputs and further fusing the spatial features with an element-wise summation.
Secondly, we model the motion dynamics of human motion via adaptive multi-grained trajectory modeling. Specifically, we summarize the temporal dynamics hierarchically with adaptive granularities according to the length of the inputs and model the trajectory evolutional directions at each granularity by gathering the outputs of BSMEs at each level along the time axis and applying temporal convolution over the trajectory.
Thirdly, we aggregate multi-granularity information by element-wise summation for accurate predictions. Finally, to get a lightweight model and avoid error accumulations, we predict future poses in a non-recursive manner.

Our main contributions are summarized as follows:
(1) we propose a brain-inspired semi-decoupled motion-sensitive encoding module, BSME, that models motion dynamics with multi-grained spatiotemporal features in a semi-decoupled manner, i.e., first decouple the process of spatial and temporal modeling via different pathways and then fuse them together. Moreover, this module provides an extra interface that can flexibly fuse any useful information and can be also used to build other spatiotemporal models.
(2) We propose a semi-decoupled multi-grained trajectory learning network, SDMTL, for predicting future human motion. In contrast to prior methods, we capture semi-decoupled spatiotemporal dynamic information via both the summarization of temporal dynamics with adaptive granularities and the modeling of temporal evolutional directions at each granularity over the whole sequence.
(3) The proposed network enables us to train a unified model for both short-term and long-term predictions instead of training different models with different settings to achieve the best performance as done in most of the prior work.
(4) Experimental results on two benchmark datasets (i.e. Human3.6M and CMU-Mocap) achieve state-of-the-art performance and evidence the effectiveness of our proposed method.

\section{Related work}

\subsection{Human motion prediction}
Most of the existing methods either implicitly modeled multi-granularity information via fixed modes, including the increased temporal receptive field in CNN models \cite{convseq2seq} and the memory mechanism in RNN models \cite{bigan,srnnap,ltowards,Ntmap}, or focused on a single granularity modeling such as the global time scale \cite{ste,LTD}. For example, the RNN models captured motion dynamics using LSTM (Long-Short Term Memory) \cite{rnmhy} or GRU (Gated Recurrent Units) \cite{guifew,GuiAdversarial} cells, and therefore their multi-granularity modeling largely relies on the memory mechanism inherently in RNNs, which suffers from inflexibly
modeling ability with various granularities especially at a coarse granularity,
leading to their poor performance.
Therefore, these models were far from enough to model the nature of human movement at multi-granularity and ignored capturing the temporal evolutional directions at each granularity over the whole sequence.

Rare work modeled motion dynamics of human motion at multi-granularity \cite{tprnn}. In \cite{tprnn}, the authors modeled multi-granularity information at multi-levels using LSTM hierarchically and augmented their features by concatenating the information of multi-granularity from the corresponding level. However, the RNN-based model ignored a part of spatial correlations among joints of the human body, since the spatial features of the pose are greatly varied with time, making it hard to achieve accurate predictions especially at a longer horizon.

\subsection{Multi-granularity analysis}
There exist some techniques on modeling multi-granularity information for other visual tasks such as action recognition, which can be roughly classified into two categories, including RNN-based methods and CNN-based methods.
The RNN-based methods modeled the temporal information of human motion at multi-granularity using LSTMs hierarchically \cite{tprnn} or sliding LSTMs over the sequence \cite{lee2017ensemble}. In \cite{lee2017ensemble}, the authors proposed ensemble Temporal Sliding LSTM (TS-LSTM) networks to model multi-granularity information, only containing short-term, medium-term, and long-term dependencies, of skeletal sequence for recognition and ignored temporal dependencies at more granularities and losed a part of spatial features of the human body, making it hard to model various movements of humans.
The CNN-based methods modeled multi-granularity information by manually setting the kernels with different sizes \cite{timeception}, manually grouping sequence with various sizes of windows \cite{trrvideo}, dilated convolutions with different dilation rates \cite{liglobal,ssnet}, convolution with the temporal shift operation \cite{tsm}, and temporal pyramid network \cite{tpnet}.

Although great success has been made on modeling multi-granularity information for other visual tasks, they heavily depend on manual parameters such as the window size of time scales, making it difficult to deploy in real diverse scenes. In contrast, our proposed SDMTL models motion dynamics with multi-grained trajectory modeling without depending on any manual parameters.

\section{Methodology}
We here introduce the proposed network for predicting 3D human motion in detail. Our model aims to predict a series of future poses conditioning on a stream of history poses. Let ${\emph {\textbf {X}}_{t - T + 1:t}} = [{\emph {\textbf {x}}_{t - T + 1}},{\emph {\textbf {x}}_{t - T + 2}}, \cdots ,{\emph {\textbf {x}}_t}]$ be the input sequence with length $T$, where $t$ is the current time-step and ${\emph {\textbf {x}}_i}(i = t - T + 1,t - T + 2, \cdots ,t)$ is the $i$-th pose of the input sequence. Here, each pose is parameterized by a group of 3D joint coordinates and the size of each pose is ($N_j \times 3$), where $N_j$ is the number of joints and 3 is the coordinate dimension of each joint. Similarly, the output future sequence is represented by ${\emph {\textbf {X}}_{t + 1:t + T'}}$ with length $T'$.

\subsection{Key components}
Spatial and motional features are two keys to describing motion sequence for predicting future dynamics. Below, we introduce several components used to build our sequential model for capturing spatial and motional features of human motion.

\subsubsection{Spatial Encoding module (SE)}
\begin{figure}[t]
\centering
\includegraphics[width=0.92\columnwidth,height=1.4in, trim = 2mm 50mm 3mm 42mm, clip=true]{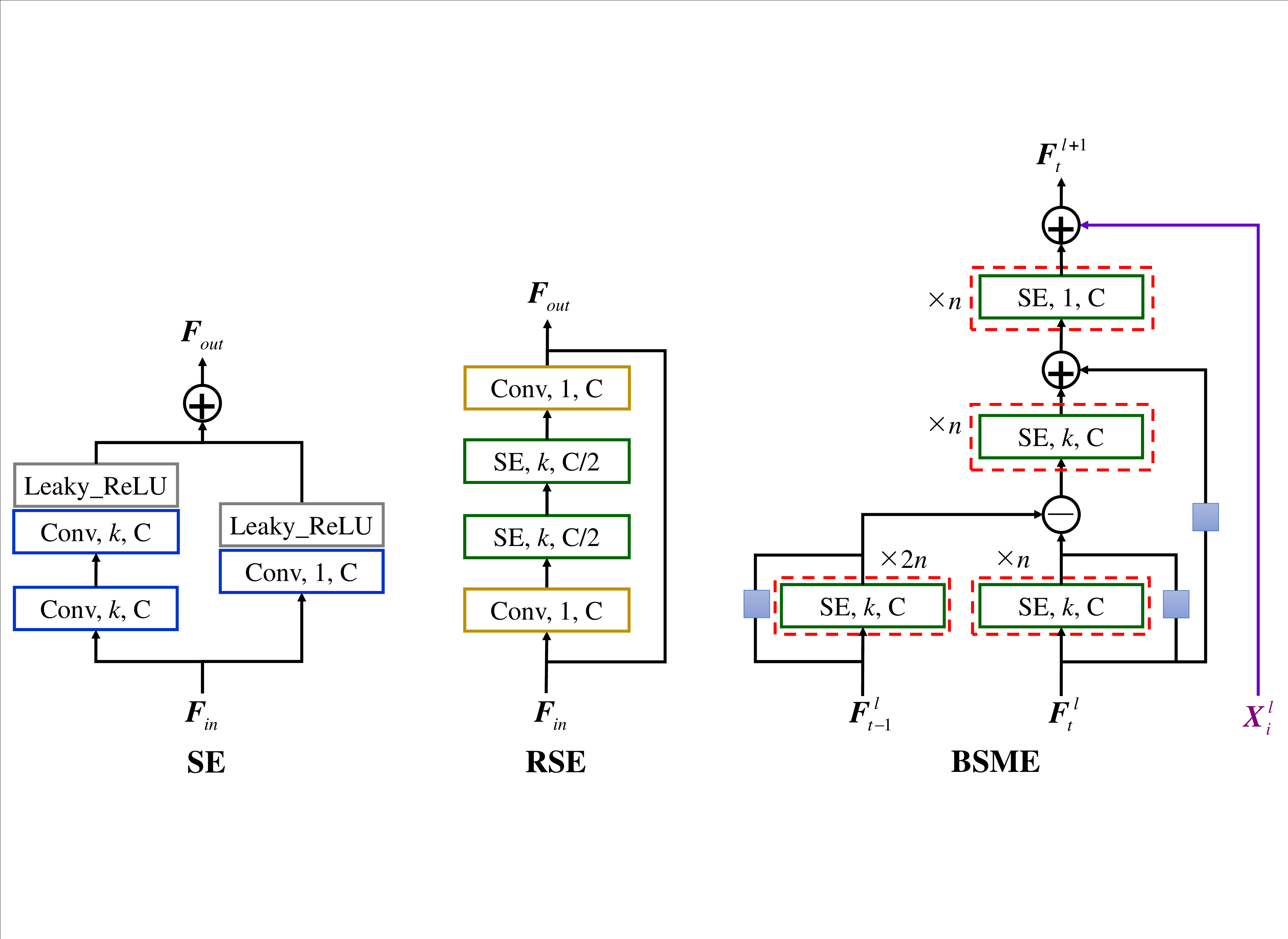}
\caption{The implementation of key components, where $C$ is the number of output channels, $k$ is the kernel size of convolutional layers, $(Conv, k, C)$ denotes a convolutional layer, $(SE, k ,C)$ denotes the SE module, ``$\times n$'' denotes ``stacked $n$ times'',
\textcircled{-} denotes element-wise subtraction, and \textcircled{+} denotes element-wise summation.} 
\label{keycompos}
\end{figure}

Inspired by the architecture of residual block \cite{he2016deep}, we propose a new SE module to model the spatial correlations among joints of the human body, and the detailed architecture of SE is shown in Figure \ref{keycompos}. There are two major differences from the commonly used residual block: (1) we discard the batch normalization (BN) layer. Different from traditional classification tasks, the data distributions of human movements should be different, and using BN layers will degrade the performance of our predictive model. Therefore, we did not use the BN layers. (2) We add a $1 \times 1$ Conv layer during the skip connection. This enables the SE to learn the pixel-level features from the inputs and then augments the output features with the pixel-level features.

\subsubsection{Residual Spatial Encoding module (RSE)}
Inspired by the squeezed channel scheme of RMB (residual multiplicative block) \cite{vpn}, we build an RSE module using SEs, and the detailed implementation is shown in Figure \ref{keycompos}. We have empirically shown that the squeezed channel scheme in RSE, i.e., first reduce the channel number of followed two stacked SEs by half and then restore the channel number to that of the inputs, can not only promote the computational efficiency but also improve the performance of the module to some extent.

\subsubsection{Brain-inspired Semi-decoupled Motion-sensitive Encoding module (BSME)}
As is shown in Figure \ref{keycompos}, BSME is built with SEs to model motion dynamics of two adjacent inputs and provides an extra interface that can further fuse other information via a shortcut and element-wise summation according to the specific need, where $\emph {\textbf {F}}^l_t$ denotes the tensor of the $l$-th level at the $t$-th time-step, and $\emph {\textbf {X}}^l_i$ denotes the information that can be fused.
Compared with the current pose, the spatial structure of early poses is increasing varying with time, and thus the spatial modeling of the previous pose is more complex than that of the current pose. Therefore, we use more SEs to encode the spatial features of previous input ${\emph {\textbf {F}}^l_{t-1}}$ and use less SEs to encode the spatial features of the current input ${{\emph {\textbf F}}^l_t}$.

BSME is built based on two insights: (1) Semi-decoupled motion-sensitive features.
Inspired by the mechanisms of the human brain that receives spatial and temporal information via different pathways respectively, we model the spatiotemporal representation in a semi-decoupled manner, i.e., first decouple the process of spatial and temporal modeling and then fuse them together.
Specifically, we first model the spatial features of each input using stacked SEs and then obtain the motion features of the adjacent inputs by element-wise subtraction, and the motion features are
further process using another stacked SEs. Finally, the semi-decoupled spatiotemporal features can be obtained by fusing the motion features with spatial features via short connection and element-wise summation. During the subtraction, the static features are eliminated and the remained features are sensitive to the human motion, therefore, the proposed BSME is motion-sensitive.
(2) Multi-grained features. The blue boxes in Figure \ref{keycompos} denote the $1 \times 1$ Conv layers without activated function, enabling the network to obtain the axis-level features. With the multi-level short connections in BSME, we can get multi-grained features as the output by gradually fusing these axis-level features. Moreover, we empirically show that these multi-grained features can greatly help to capture rich motion dynamics for predictions.

\subsection{Architecture of SDMTL}
\begin{figure*}[t]
\centering  
\includegraphics[width=1.65\columnwidth,height=2.9in, trim =50mm 6mm 45mm 12mm, clip=true]{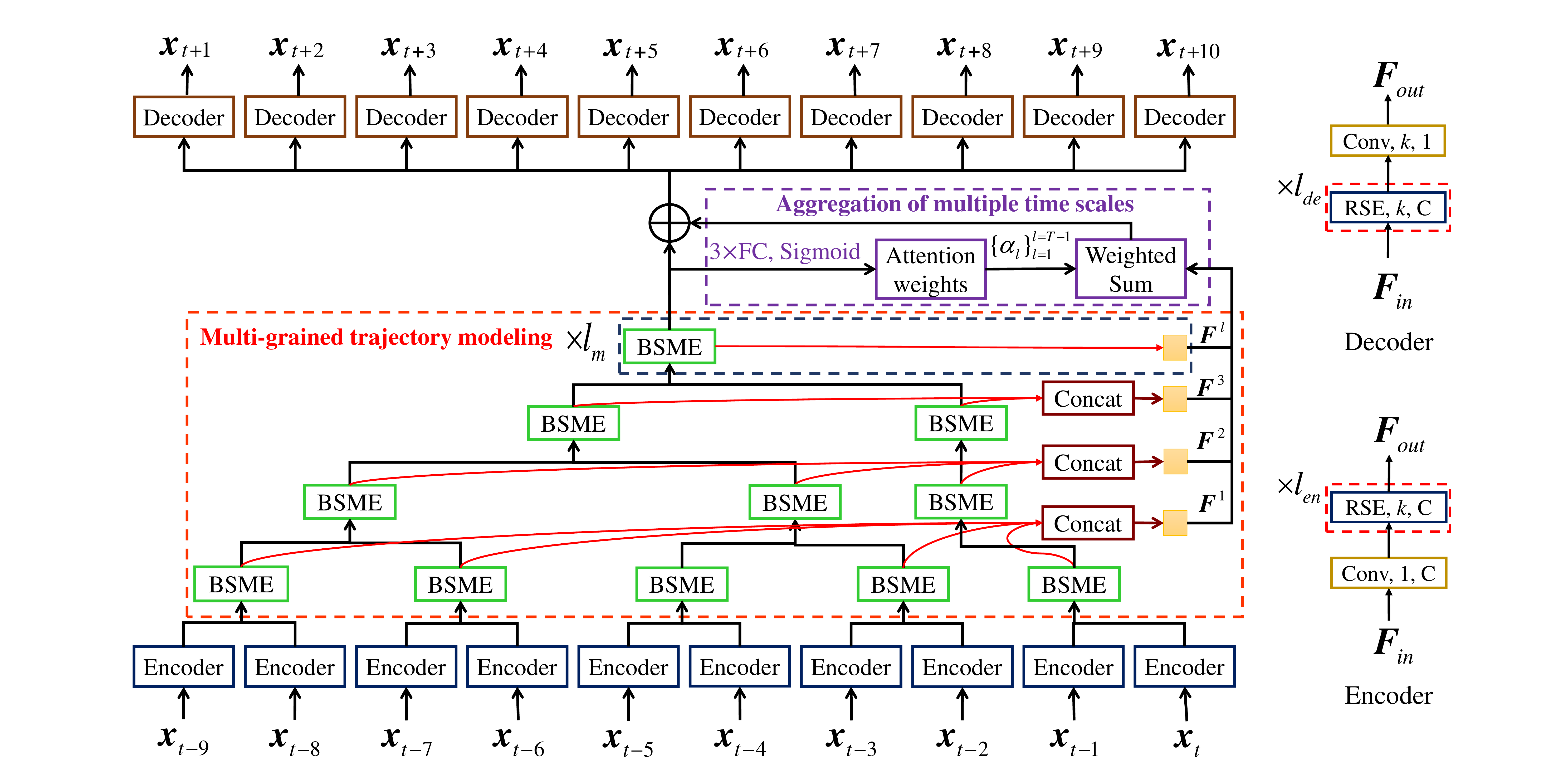} 
\caption{Overall architecture. Here, we give an illustration of our SDMTL that predicts 10 future poses conditioning on 10 historical poses, where the orange boxes denote the $k \times k$ Conv layers followed by a Leaky ReLU layer.}
\label{amta}
\end{figure*}

The overall architecture is shown in Figure \ref{amta}, mainly including three parts: Encoder/Decoder, multi-grained trajectory modeling, and aggregation of multi-granularity.

\subsubsection{Encoder/Decoder}
The encoder is to encode spatial correlations among joints of the human body, and the decoder is to decode spatial information of the human body. Therefore, the process of encoder and decoder is similar, and the decoder can be regarded as the inverse process of the encoder. In this paper, we propose an encoder and decoder with similar architecture, as is shown at the right of Figure \ref{amta}. Specifically, for the encoder, a $1 \times 1$ Conv layer is first applied to model the axis-level features of joints, and then stacking $l_{en}$ SEs enlarges the spatial receptive field to model the spatial correlations among joints of the human body; for the decoder, stacking $l_{de}$ SEs reconstructs the spatial features of the future pose,
and we apply another Conv layer and set the number of output channels to 1 since we focus on predicting a single pose at one time.

%

\subsubsection{Multi-grained trajectory modeling}
As is shown in Figure \ref{amta}, the multi-grained trajectory modeling is built with BSMEs at multi-levels to summarize the temporal information at multi-granularity and gathering the outputs of each scale to model temporal evolutional directions of their trajectories. Therefore, our multi-grained trajectory modeling both contains
temporal dynamic summarization with adaptive granularities and temporal evolutional trajectory modeling.

For temporal dynamic summarization with adaptive granularities, we focus on summarizing temporal dynamics at multi-granularity using BSMEs on multi-levels and model each granularity at each level.
In the $l$-th level, we model the temporal information at scale $2^l$.
The BSMEs in the same level share parameters, which can build a lightweight representation for each granularity and also improve the performance potentially in the mean time. At the $\left\lceil {{{\log }_2}T} \right\rceil$-th level (``$\left\lceil  \cdot  \right\rceil $'' denotes a ceil function), we can model the global temporal information of previous $T$ poses.
Moreover, we empirically show that more levels can further improve the performance of the model, and therefore we stack more levels with the same inputs at the $(\left\lceil {{{\log }_2}T} \right\rceil+1)$-th level to the $(\left\lceil {{{\log }_2}T} \right\rceil+{l_m})$-th level. Here, we empirically show that setting the total number of levels to ($T$-1) works the best. Therefore, the total number of time scales is adaptive to the length of the input $T$.

The information flow of the extra interface of BSMEs is hidden in Figure \ref{amta}. In this paper, let $\{\emph {\textbf {F}}^0_i\}$ ($i=1,2,\cdots,T$) be the output of the $i$-th encoder, we initialize the input of the extra interface to $\{\emph {\textbf {F}}^0_i\}$ as $\emph {\textbf {X}}^l_j = \emph {\textbf {F}}^0_k$ ($k={{2^l}j}$, $j=1,2,\cdots,\left\lceil {\frac{T}{{{2^{l - 1}}}}} \right\rceil $), where $\emph {\textbf {X}}^l_j$ denotes the input of the extra interface of the $j$-th BSME in the $l$-th level.
In this way, we can promote the information flow among different levels to better capture multi-granularity features and further enhance the captured semi-decoupled spatiotemporal features by gradually fusing the spatial features from the low layers.

For temporal evolutional trajectory modeling, we focus on capturing the information of temporal evolutional directions at each granularity. For this, we gather the outputs of BSMEs at each level by concatenating along the time axis and then apply one Conv layer to conveniently capture the evolutional directions of this granularity since the temporal dimension is set as the channel of the input tensor.

To sum up, with the elegant design in both BSME and multi-grained trajectory modeling, the spatiotemporal information flows from the low level to a higher level in a repeated semi-decoupled manner.

\subsubsection{Aggregation of multi-granularity}
Different human motion reflects different multi-granularity, and therefore it is necessary to aggregate the information with proper granularities. As is shown in Figure \ref{amta}, stacking 3 FC layers with sigmoid as the activation function is applied to the top level of multi-grained trajectory modeling to learn a group of weights for each level, denoting by $\{ {\alpha _l}\} _{l = 1}^{l = T - 1}$. Given the features at the $l$-th level as $\emph{\textbf{F}}^l$ as is denoted in Figure \ref{amta}, the multi-granularity information can be aggregated by weighted summation as equation \ref{eqn1}.

\begin{equation}
{\emph{\textbf{F}}} = \sum\limits_{l = 1}^{l = T - 1} {{\alpha _l}} {{\emph{\textbf{F}}}^l}
\label{eqn1}
\end{equation}

\subsection{Loss}
Inspired by \cite{agvnet}, we use the Temporal Weighted Mean Per Joints Position Error (TW-MPJPE) loss shown in the equation \ref{eqn2} to optimize our model, improving the short-term predictions by paying increasing attention to the early time-steps. Different from \cite{agvnet}, the temporal weights of our loss are decreasing nonlinearly with the time since the difficulty of predictions at different time-steps should be nonlinear.

\begin{equation}
\begin{array}{l}
L{\rm{ = }}\frac{1}{{{N_j}}}{\sum\limits_{i = 1}^{T'} {{w_i}\sum\limits_{j = 1}^{{N_j}} {\left\| {J_j^i - \widehat J_j^i} \right\|} } ^2}\\
{w_i} = \frac{{{e^{ - \alpha i}}}}{{\sum\limits_{t = 1}^{T'} {{e^{ - \alpha t}}} }}
\end{array}
\label{eqn2}
\end{equation}
Where ${T'}$ is the number of future poses, $N_j$ is the number of joints, $J_j^i$ and $\widehat J_j^i$ denotes the groundtruth and prediction of the $j$-th joint at the $i$-th time-step, respectively.

\section{Experiments}
In this section, we first describe the datasets used in this paper and the implementation details of SDMTL, and then we briefly introduce the baselines for comparison. Next, we evaluate the performance of our model with both quantitative and qualitative analyses. Finally, we conduct ablative experiments to analyze the main contributions of SDMTL.

\subsection{Datasets and Implementation Details}
\subsubsection{Datasets}
(1) Human3.6M \cite{h36m} (H3.6M) is the most widely used dataset for 3D human motion prediction, containing 7 actors performing 15 daily activities such as walking and smoking. (2) CMU mocap dataset (CMU-Mocap)\footnote{\url{http://mocap.cs.cmu.edu/}}: consistent with the baselines, eight actions are selected for evaluations, including running, basketball, etc.

\subsubsection{Implementation Details}
Consistent with baselines, we adopt the same training set, validating set, and testing set for experiments and apply the same data processing. All models are implemented with TensorFlow. Adam optimizer is used to optimize our models on one GTX 1080Ti GPU by minimizing our loss, and the learning rate is initialized to 0.0001. The number of output channels (i.e. $C$) is set to 64 and the kernel size (i.e. $k$) is set to 3. In experiments, we empirically show that these hyperparameters work the best, i.e. $l_{en}$ =1, ${l_{de}}$ =1 and $\alpha$=0.3. The skeletal representation used in this paper is the same as that in \cite{pisepp}. We use MPJPE in millimeter as our metric. We predict 10 future poses (400 milliseconds) for short-term prediction and predict 25 future poses (1 second) for long-term prediction, conditioning on 10 historical poses.
All experimental settings are consistent with the baselines.
More details and experimental results can be found in the supplemental materials.

\subsection{Baselines}
We compare our methods with several recent methods such as RRNN \cite{hmprnn}, ConvSeq \cite{convseq2seq}, LTD \cite{LTD}. RRNN and ConvSeq are two models that implicitly model multi-granularity information via fixed modes. LTD focuses on modeling a single granularity in the trajectory space via DCT, i.e., a global time scale, and it also achieves state-of-the-art performance.

\subsection{Comparison with baselines}

\begin{table*}[!t]
\scriptsize
\centering  
\begin{tabular}{p{1.41cm}p{0.15cm}p{0.15cm}p{0.25cm}p{0.25cm}p{0.28cm}p{0.4cm}|
p{0.15cm}p{0.15cm}p{0.25cm}p{0.25cm}p{0.28cm}p{0.4cm}|
p{0.15cm}p{0.15cm}p{0.25cm}p{0.25cm}p{0.28cm}p{0.4cm}|
p{0.15cm}p{0.15cm}p{0.25cm}p{0.25cm}p{0.28cm}p{0.4cm}}
\hline
\multirow{2}{*}{Milliseconds}& \multicolumn{6}{c|}{Walking} & \multicolumn{6}{c|}{Eating}& \multicolumn{6}{c|}{Smoking} & \multicolumn{6}{c}{Discussion}\\
\cline{2-25} & 80 &160 & 320 &400 & 560 &1000 & 80 &160 & 320 &400 & 560 &1000&
 80 &160 & 320 &400 & 560 &1000 & 80 &160 & 320 &400& 560 &1000\\
\hline

RRNN &23.8 &40.4& 62.9& 70.9 &73.8 &86.7& 17.6& 34.7& 71.9& 87.7  & 101.3& 119.7& 19.7& 36.6& 61.8& 73.9& 85.0 &118.5& 31.7& 61.3& 96.0& 103.5 & 120.7& 147.6 \\
ConvSeq  &17.1 &31.2&53.8&61.5  & 59.2& 71.3&13.7&25.9&52.5&63.3& 66.5& 85.4&11.1&21.0&33.4&38.3& 42.0& 67.9& 18.9&39.3&67.7&75.7 & 84.1& 116.9 \\
\hline
 LTD-short &8.9 &15.7 & 29.2& 33.4 &{--}&{--}& 8.8 &18.9&	39.4&47.2&{--}&{--}&7.8 &14.9&25.3&{\bf 28.7}&{--}&{--}& 9.8 &22.1 &39.6&44.1 &{--}&{--}\\
LTD-long &11.2 &19.1 &32.6 &37.6 &42.2 &{\bf 51.3}& 9.0 &18.4 &37.2 &45.1 &56.5 &{\bf 68.6} &8.3 &15.9 &26.3 &29.6 &{\bf 32.3} &60.5 &12.0 &24.9 &41.3 &46.3 &70.4 &{\bf 103.5} \\
\hline
SDMTL-short&8.3 &16.6 & 29.5 &34.4 & {--}&{--}& 8.4 &16.9 & 36.0 & 44.4 & {--}&{--}& {\bf 6.3} & {\bf 13.2} & 25.4 & 30.2 & {--}&{--}&{\bf 7.7} &{\bf 20.4} &{\bf 40.4} &47.0 & {--}&{--}\\
SDMTL-long &{\bf 7.6} &{\bf 14.8} & {\bf 25.9} & {\bf 29.1} &{\bf 36.3} & 53.5 &{\bf 8.2} &{\bf 16.4} &{\bf 33.8} &{\bf 42.4} &{\bf 53.9} &68.8 & 6.4 &13.4 &{\bf 24.3} &{ 29.3} & 33.5 &{\bf 58.4} &7.8 & 20.6 & 40.7 & {\bf 45.9} & {\bf 69.5} & 104.4 \\
\hline
\hline
\multirow{2}{*}{Milliseconds}& \multicolumn{6}{c|}{Directions} & \multicolumn{6}{c|}{Greeting}& \multicolumn{6}{c|}{Phoning} & \multicolumn{6}{c}{Posing}\\
 \cline{2-25}& 80 &160 & 320 &400& 560 &1000 & 80 &160 & 320 &400& 560 &1000&
  80 &160 & 320 &400 & 560 &1000& 80 &160 & 320 &400& 560 &1000\\
\hline
RRNN  & 36.5 &56.4& 81.5& 97.3  &{--}&{--}& 37.9& 74.1& 139.0& 158.8   &{--}&{--}&25.6& 44.4& 74.0& 84.2&{--}&{--}& 27.9& 54.7& 131.3& 160.8  &{--}&{--}\\
ConvSeq  & 22.0&37.2 &59.6& 73.4 &{--}&{--} &24.5 &46.2 &90.0& 103.1&{--}&{--}& 17.2& 29.7& 53.4 &61.3&{--}&{--}& 16.1& 35.6& 86.2& 105.6 &{--}&{--}\\
\hline
LTD-short &12.6 &24.4 &48.2 &58.4 &{--}&{--}&14.5 &30.5 &74.2 	&89.0&{--}&{--}&11.5 	&20.2 &37.9 &43.2 &{--}&{--}&9.4 &23.9 &66.2 &82.9 &{--}&{--}\\
LTD-long &13.5 &25.7 &50.5 &62.0 &{\bf 85.8} &109.3 &15.6 &30.9 &68.5 &82.3 &91.8 &{\bf 87.4} &13.0 &21.6 &41.2 &47.4 &65.0 &113.6 &11.1 &26.3 &69.9 &87.3 &113.4 &220.6\\
\hline
SDMTL-short &{\bf 8.8} &{\bf 20.4} &{\bf 46.5} &{\bf 58.1} &{--}&{--}&{\bf 11.2} &{\bf 25.1} & 65.6 & 81.9&{--}&{--} & {\bf 10.2} & 18.9 & {\bf 34.2} &{\bf 40.0} &{--}&{--}&7.2 &21.3 &{\bf 63.4} &{\bf 80.7} &{--}&{--}\\
SDMTL-long& 9.8 & 23.4 & 53.8 & 67.0 & 88.3 &{\bf 107.9} & 11.7 & 25.3 &{\bf 61.9} &{\bf 75.0} & {\bf 88.7} & 89.0 &10.5 &{\bf 18.5} & 37.2 & 43.1 &{\bf 60.8} &{\bf 112.3} &{\bf 6.8} &{\bf 20.5} & 64.0 & 82.4 &{\bf 107.2} &{\bf 204.7} \\
\hline
\hline
\multirow{2}{*}{Milliseconds}& \multicolumn{6}{c|}{Purchases} & \multicolumn{6}{c|}{Sitting}& \multicolumn{6}{c|}{Sitting Down} & \multicolumn{6}{c}{Taking Photo}\\
\cline{2-25} & 80 &160 & 320 &400 & 560 &1000 & 80 &160 & 320 &400 & 560 &1000&
 80 &160 & 320 &400 & 560 &1000& 80 &160 & 320 &400& 560 &1000\\
\hline
RRNN  & 40.8& 71.8& 104.2& 109.8   &{--}&{--}&34.5& 69.9& 126.3 &141.6&{--}&{--}& 28.6& 55.3& 101.6& 118.9 &{--}&{--}& 23.6 &47.4& 94.0& 112.7 &{--}&{--} \\
ConvSeq & 29.4& 54.9& 82.2& 93.0 &{--}&{--}&19.8 &42.4& 77.0& 88.4 &{--}&{--}& 17.1& 34.9& 66.3& 77.7&{--}&{--}& 14.0& 27.2& 53.8& 66.2  &{--}&{--}\\
\hline
LTD-short &   19.6 &38.5 &64.4 &72.2 &{--}&{--}&10.7 &24.5 &50.6 &{\bf 62.0} &{--}&{--}&11.4 &27.6 &56.4 &67.6 &{--}&{--}&6.8 &15.2 &38.2 &49.6 &{--}&{--}\\
LTD-long &21.4 &41.9 &64.0 &72.9 &94.3 & 130.4 & 11.5 &27.2 &54.0 &{ 64.1} &{\bf 79.6} &{\bf 114.9} &12.7 &28.6 &55.1 &64.2 &82.6 &140.1 &7.8 &16.6 &39.9 &50.4 &68.9 &{\bf 87.1} \\
\hline
SDMTL-short&{\bf 16.6} &{\bf 35.0} &{\bf 59.9} &{\bf 66.4} &{--}&{--}&{\bf 8.4} &22.8 &52.7 &{ 65.4}  &{--}&{--}&9.9 &27.5 &55.0 &63.4 &{--}&{--}&6.1 &14.5 &37.0 &47.6 &{--}&{--}\\
SDMTL-long&  18.4 &38.8 &61.1 &68.2 &{\bf 80.9} &{\bf 113.6} & 8.7 &{\bf 22.2} &{\bf 52.2} & 65.5 &83.9 &115.5 &{\bf 9.3} &{\bf 23.8} &{\bf 50.6} &{\bf 60.9} &{\bf 77.7} &{\bf 118.9} &{\bf 6.0} &{\bf 14.0} &{\bf 36.1} &{\bf 47.0} &{\bf 67.1} & 91.1 \\
\hline
\hline
\multirow{2}{*}{Milliseconds}& \multicolumn{6}{c|}{Waiting} & \multicolumn{6}{c|}{Walking Dog}& \multicolumn{6}{c|}{Walking Together} & \multicolumn{6}{c}{Average}\\
 \cline{2-25}& 80 &160 & 320 &400 & 560 &1000 & 80 &160 & 320 &400& 560 &1000 &
  80 &160 & 320 &400 & 560 &1000& 80 &160 & 320 &400& 560 &1000\\
\hline
RRNN & 29.5& 60.5& 119.9& 140.6&{--}&{--}& 60.5& 101.9& 160.8& 188.3&{--}&{--}& 23.5& 45.0& 71.3& 82.8&{--}&{--}&  30.8& 57.0& 99.8& 115.5  &{--}&{--}\\
ConvSeq &17.9& 36.5& 74.9& 90.7&{--}&{--}& 40.6& 74.7& 116.6& 138.7&{--}&{--}& 15.0& 29.9& 54.3& 65.8&{--}&{--}& 19.6& 37.8& 68.1& 80.2  &{--}&{--}\\
\hline
LTD-short & 9.5 &22.0 	&57.5 &73.9 &{--}&{--}&32.2 &58.0 &102.2 &122.7 &{--}&{--}&8.9 &18.4 &35.3 &44.3 &{--}&{--}&12.1 &25.0 &51.0 &61.3 &{--}&{--}\\
LTD-long &10.1 &23.0 &59.4 &75.7 &100.9 &167.6 &34.9 &61.1 &102.1 &120.4&{\bf 136.6} &{\bf 174.3} &10.2 &20.6 &36.1 &44.3 &57.0 &85.0 &13.5 &26.8 &51.9 &62.0 &78.5 &114.3 \\
\hline
SDMTL-short&{\bf 7.1} & 19.3 & 50.5 & 64.9 &{--}&{--}&{\bf 20.6} &{\bf 52.2} &{\bf 94.4} &{\bf 111.2} &{--}&{--}&{\bf 6.5} &{\bf 13.9} &{\bf 30.5} &39.8 &{--}&{--}&{\bf 9.5} &{\bf 22.5} &48.1 & 58.4 &{--}&{--}\\
SDMTL-long & 7.5 &{\bf 19.0} &{\bf 46.8} &{\bf 58.3} &{\bf 81.4} &{\bf 159.2} & 21.0 & 54.9 & 100.4 & 119.8 & 137.7 & 181.5 &7.1 & 15.4 & 31.5 &{\bf 39.7} &{\bf 50.8} &{\bf 81.1} & 9.8 & 22.7 & {\bf 48.0} & {\bf 58.2} & {\bf 74.5} &{\bf 110.7} \\
\hline
\end{tabular}
\caption{Short and long-term prediction on H3.6M. The 3D errors for 4 activities of ``RRNN''
and ``ConvSeq'' models for long-term prediction are provided in ``LTD''. ``*-short'' and ``*-long'' denote the short-term and long-term predictive model of method ``*'', respectively. }
\label{rh36m}
\end{table*}


\begin{table*}[!t]
\scriptsize
\begin{center}
\begin{tabular}{p{1.42cm}p{0.1cm}p{0.2cm}p{0.2cm}p{0.2cm}p{0.3cm}p{0.4cm}|p{0.1cm}p{0.2cm}p{0.2cm}p{0.3cm}p{0.2cm}p{0.4cm}|
p{0.1cm}p{0.2cm}p{0.2cm}p{0.2cm}p{0.3cm}p{0.4cm}|p{0.1cm}p{0.2cm}p{0.2cm}p{0.2cm}p{0.3cm}p{0.4cm}}
\hline
\multirow{2}{*}{Milliseconds}& \multicolumn{6}{c|}{Basketball} & \multicolumn{6}{c|}{Basketball Signal}& \multicolumn{6}{c|}{Directing Traffic}& \multicolumn{6}{c}{Jumping}\\
\cline{2-25} & 80 &160 & 320 &400 &560 & 1000 & 80 &160 & 320 &400 &560& 1000 & 80 &160 & 320 &400 &560 & 1000 & 80 &160 & 320 &400&560 & 1000\\
\hline
LTD-short &14.0& 25.4& 49.6& 61.4& {--}&{--}& 3.5& 6.1& 11.7& 15.2& {--}&{--} & 7.4& 15.1& 31.7& 42.2& {--}&{--}& 16.9& 34.4& 76.3& 96.8& {--}&{--}\\
LTD-long& 13.7 &24.2 &46.7 &58.0 &77.4 &{\bf 99.9}&3.4 &6.0 &{\bf 12.7} &{\bf 16.5} &{\bf 25.3} &{\bf 53.3} &7.4 &14.5 &32.7 &43.6 &70.3 &149.4& 16.6 &33.6 &77.5 &99.4 &131.4 &{\bf 171.7}  \\
\hline
SDMTL-short&{\bf 10.9} &{\bf 18.9} &{\bf 39.3} &{\bf 50.0} & {--}&{--}&{\bf 2.5}&{\bf 5.5} &{ 15.3} &{ 22.1} &{--}&{--}&5.3 &{\bf 10.8} &24.6 & 33.3 & {--}&{--}&{\bf 10.7} &{\bf 22.9} &{\bf 60.5} &{\bf 82.8} & {--} &{--}\\
SDMTL-long&{\bf 10.9} & 20.2 & 40.9 &50.8 & {\bf 66.1} &110.2 & 2.9 & 6.2 & 16.4 & 23.1 & 37.4 & 71.6 &{\bf 5.1} & 10.9 &{\bf 23.2} &{\bf 30.2} & {\bf 46.1} &{\bf 104.5} &11.1 & 24.6 & 65.7 & 90.3 & {\bf 130.9} & 191.2 \\
\hline
\hline
\multirow{2}{*}{Milliseconds} & \multicolumn{6}{c}{Running}& \multicolumn{6}{c|}{Soccer}& \multicolumn{6}{c|}{Walking} & \multicolumn{6}{c}{Washwindow}\\
\cline{2-25} & 80 &160 & 320 &400 &560& 1000 & 80 &160 & 320 &400 &560& 1000 & 80 &160 & 320 &400 &560& 1000& 80 &160 & 320 &400 &560& 1000  \\
\hline
 LTD-short & 25.5& 36.7& 39.3& 39.9& {--}&{--} & 11.3& 21.5& 44.2& 55.8& {--}&{--}& 7.7& 11.8& { 19.4}& { 23.1}&{--}&{--}& 5.9& 11.9& 30.3&{40.0}&{ --}&{--}\\
 LTD-long& 24.0 &33.9 &35.3 &35.4 &{\bf 36.2} &{\bf 59.8} &11.7 &22.5 &44.7 &56.8 &82.6 &{\bf 119.0} &7.2 &11.4 &20.6 &23.7 &27.2 &32.4 &5.7 &11.9 &31.3 &41.7 &53.0 &{\bf 80.7} \\
\hline
SDMTL-short&{\bf 14.1} &{\bf 16.6} &{\bf 22.3} &{\bf 28.3} &{--}&{--}&{\bf 8.0} &{\bf 16.0} &{\bf 36.6} &{\bf 50.3} &{--}&{--}&{\bf 5.8} &{\bf 8.8} &{\bf 15.8} &{\bf 18.7} &{--}&{--}&{\bf 4.3} &{\bf 9.5} &{\bf 28.5} & 39.5 &{--}&{--}\\
SDMTL-long&14.9 &18.6 &25.1 & 30.6 & 40.4 & 68.4 & 8.1 &16.5 & {\bf 36.6} & 50.6 &{\bf 77.0} & 140.7 &6.1 & 9.0 & 17.5 & 20.0 & 26.3 & 51.9 & 4.6 & 10.1 & 29.6 &{\bf 39.2} &{\bf 50.9} &{ 82.4} \\
\hline
\end{tabular}
\end{center}
\caption{Short and long-term prediction on CMU-Mocap. }
\label{tcmu}
\end{table*}

\subsubsection{Quantitative results}
(1) Quantitative results on H3.6M. As is shown in Table \ref{rh36m}, compared with baselines, our model achieves the best performance with a unified model for both short-term and long-term prediction in most cases, demonstrating the effectiveness of our method. This contrasts with LTD \cite{LTD} that requires training different models with different settings to achieve their best performance. Moreover, on average, the performance gaps of early time-steps (i.e. \textless 500ms) between the short-term model and the long-term model of our method are significantly smaller than that of LTD, further showing the effectiveness of our method powerfully.

Our superior performance benefits from two folds.
($a$) In contrast to the coupled \cite{hmprnn,convseq2seq} or decoupled \cite{LTD} spatiotemporal modeling as done in prior work, the proposed BSME models the spatiotemporal information in a semi-decoupled manner, i.e., first decoupled the process of spatial and temporal modeling and then fuse them together. Such a semi-decoupled strategy simplifies the spatiotemporal modeling of the network that forces different components of the network to focus on modeling either spatial information or temporal information.
($b$) In contrast to the previous work that either models multi-granularity information via fixed modes or focuses on a single granularity modeling, we capture rich multi-granularity information via multi-grained trajectory learning by summarizing temporal dynamics with adaptive granularities and modeling temporal evolutional directions at each granularity.

(2) Quantitative results on CMU-Mocap. As is shown in Table \ref{tcmu}, our method consistently outperforms the baseline \cite{LTD} on average, demonstrating the effectiveness of our method again.
This thanks to our semi-decoupled spatiotemporal modeling with BSMEs and our novel network that captures rich multi-granularity information.

\subsubsection{Qualitative results}
\begin{figure}[t]
\centering
\subfigure[Taking photo]{\includegraphics[width=0.95\columnwidth,height=0.7in, trim = 65mm 18mm 55mm 15mm, clip=true]{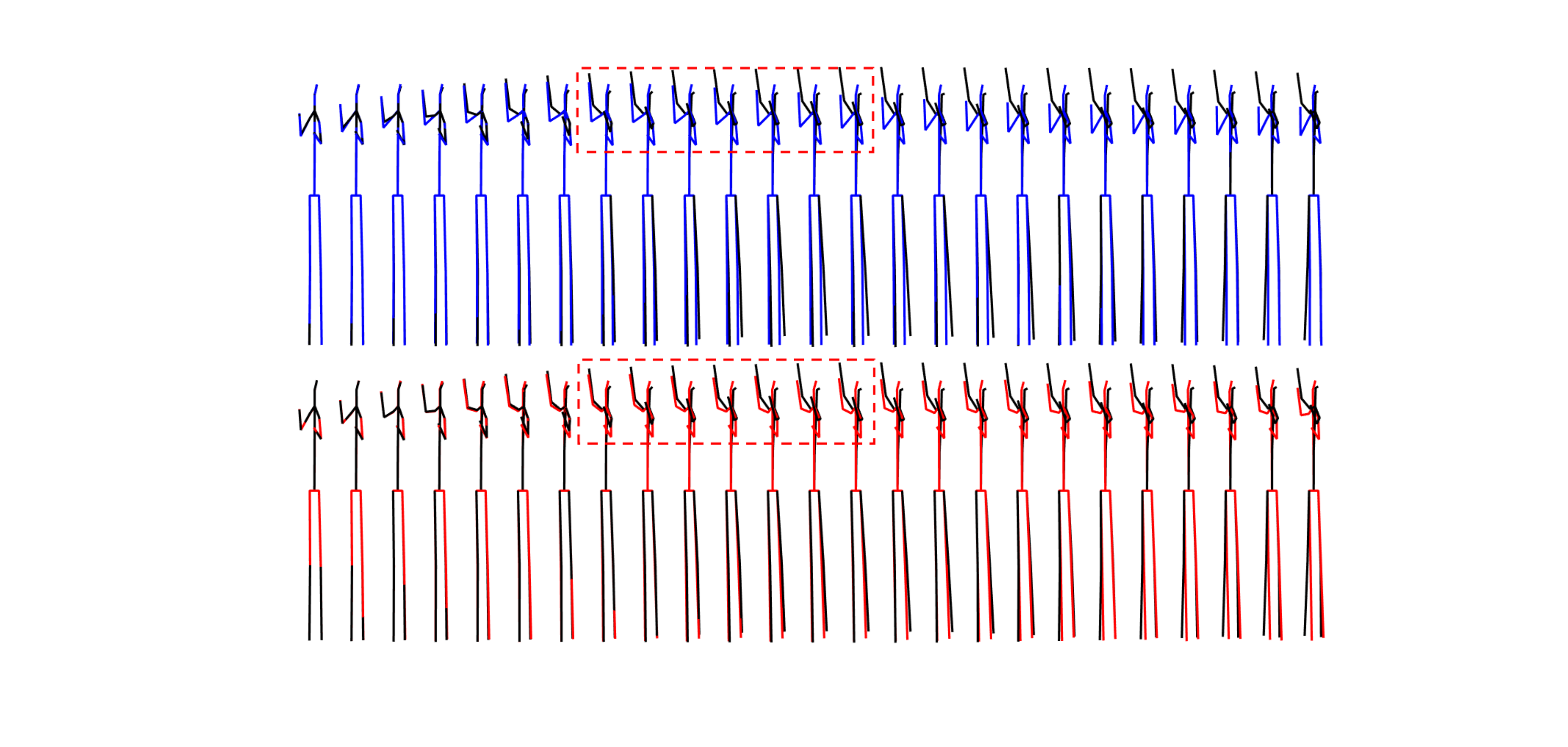}
\label{vish36m1}
}
\subfigure[Posing]{\includegraphics[width=0.95\columnwidth,height=0.7in, trim = 67mm 25mm 55mm 20mm, clip=true]{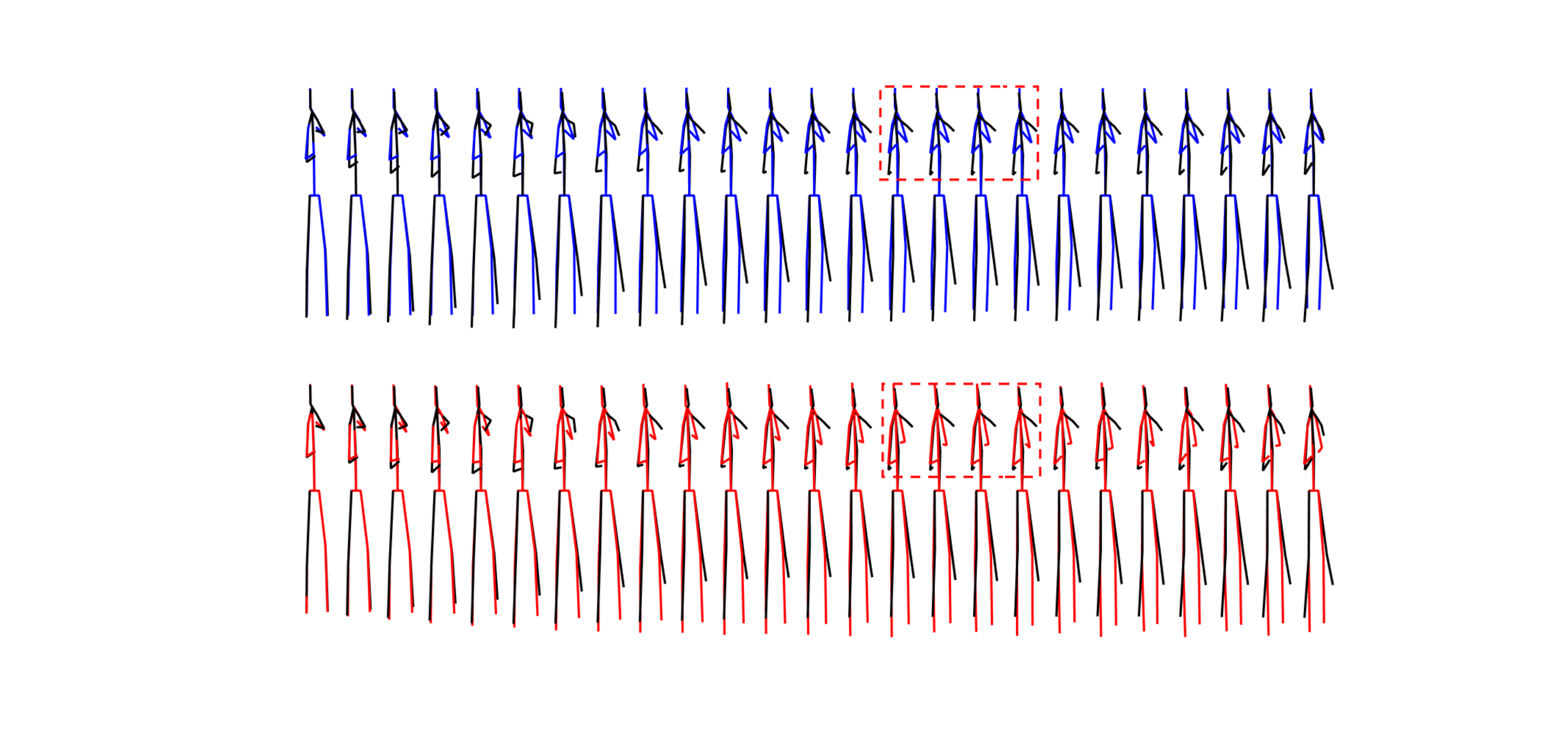}
\label{vish36m2}
}
\caption{Qualitative results on H3.6M. } 
\label{vish36m}
\end{figure}

\begin{figure}[t]
\centering
\subfigure[Directing Traffic]{\includegraphics[width=0.95\columnwidth,height=0.6in, trim = 80mm 25mm 35mm 20mm, clip=true]{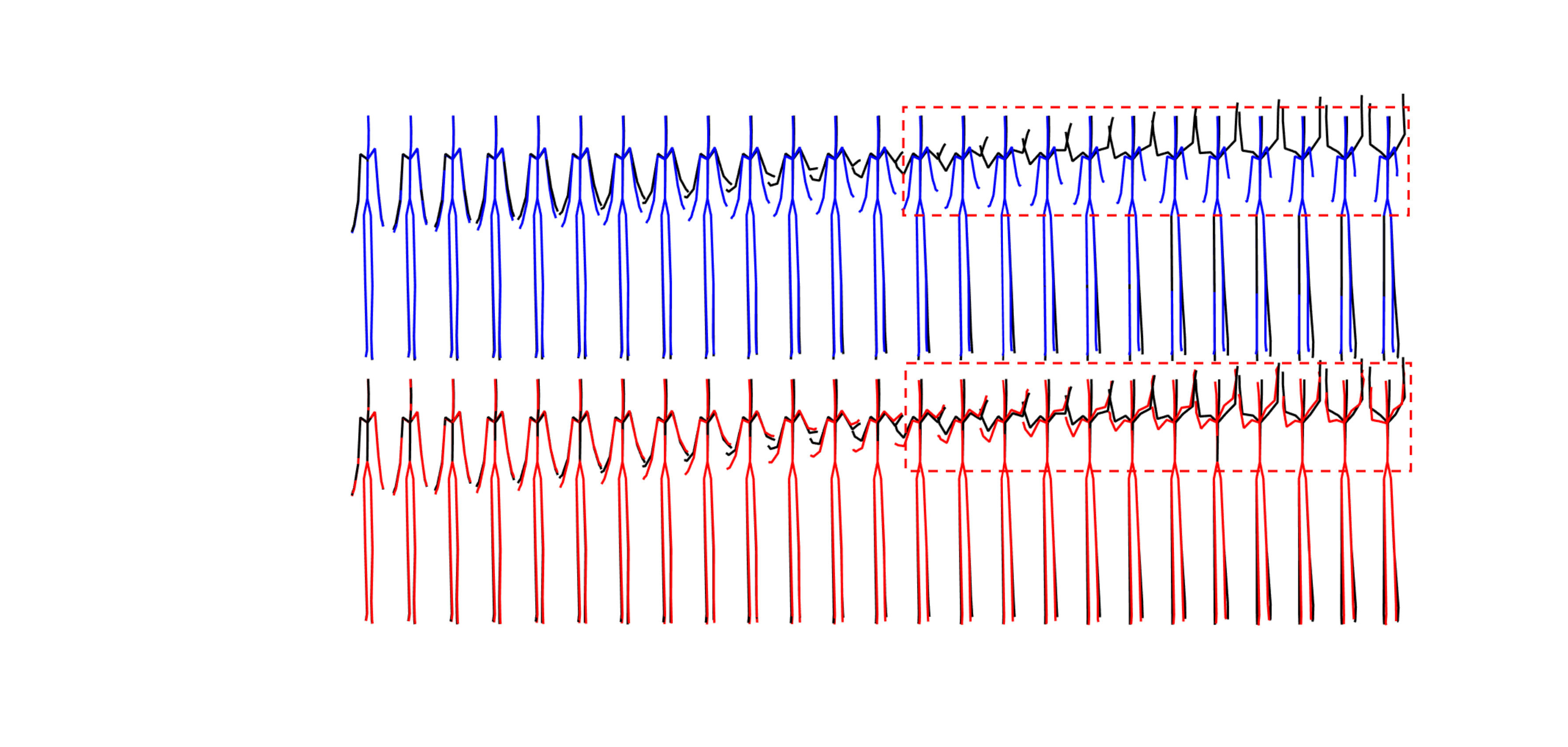}
\label{viscmu1}
}
\subfigure[Soccer]{\includegraphics[width=0.95\columnwidth,height=0.6in, trim = 47mm 26mm 37mm 28mm, clip=true]{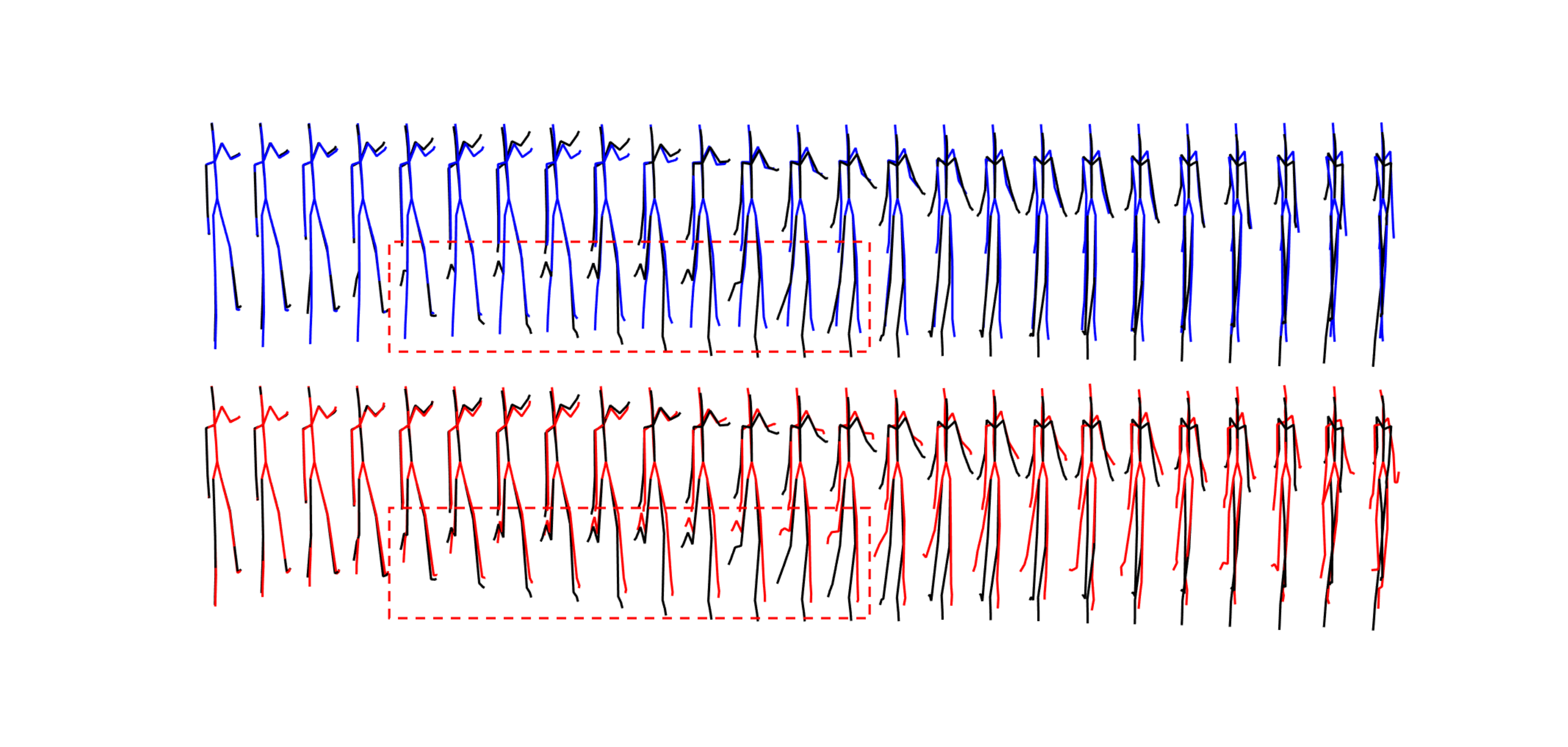}
\label{viscmu2}
}
\caption{Qualitative results on CMU-Mocap. } 
\label{viscmu}
\end{figure}

To further show the effectiveness of our method, we visualize the predictions of the long-term model frame-by-frame on H3.6M and CMU-Mocap respectively, and the results are reported Figure \ref{vish36m} and Figure \ref{viscmu}.

(1) Qualitative results on H3.6M. As is denoted in Figure \ref{vish36m}, for the right hands in both Figure \ref{vish36m1} and Figure \ref{vish36m2}, our predictions are closer to the groundtruth than that of LTD, demonstrating the effectiveness of our proposed method.
This may benefit from the advantages of our method that captures rich motion dynamics in a semi-decoupled manner via both the summarization of temporal dynamics with adaptive granularities and the modeling of temporal evolutional directions at each granularity, which contrasts to that of LTD that focuses on modeling motion dynamics at a single granularity. Therefore, we capture rich motion dynamics for accurate predictions.

(2) Qualitative results on CMU-Mocap. As is shown in Figure \ref{viscmu}, for the hands in Figure \ref{viscmu1}
 and the right legs in Figure \ref{viscmu2}, our predictions are significantly better than that of LTD, showing the effectiveness of our method again. Specifically, for the predictions of LTD, the right legs of predictive poses change slightly over time, contrasting to the groundtruth. By contrast, the right legs of our predictive poses change greatly, and the temporal evolutional law is basically consistent with that of the groundtruth, demonstrating the effectiveness of our method powerfully.
Moreover, although the predictions at the longer time-steps are slightly worse than that of LTD, as is shown in Figure \ref{viscmu2}, the generative poses are still reasonable. The main reasons for our superior performance are two-fold: ($a$) the proposed BSME captures semi-decoupled spatiotemporal features and it is sensitive to the motion; ($b$) we capture rich motion dynamics of human movement via multi-grained trajectory learning.
This is different from that of LTD that focuses on modeling motion dynamics at a single granularity and ignores the nature of human motion at more granularities. Therefore, we can better capture motion dynamics for predicting the complex human motion.

\begin{table}[!t]
\scriptsize
\centering
\begin{tabular}{ccccccc}
\hline
\multirow{2}{*}{Milliseconds}&\multicolumn{6}{c}{Evaluation of rich multiple time scale modeling} \\
\cline{2-7}
{} & 80 &160 & 320 &400 &560 & 1000\\
 \hline
{w/o TED} &9.9 & 22.8 & 48.5 & 58.8 & 75.3 & 111.4 \\
{w/o AMG}&10.5 & 23.2 & 48.6 & 58.6 & 74.9 & 111.8 \\
 \hline
{SDMTL (Ours)} &{\bf 9.8} &{\bf 22.7} & {\bf 48.0} & {\bf 58.2} & {\bf 74.5}&{\bf 110.7} \\

 \hline
\end{tabular}
\caption{Results of ablative experiments on H3.6M. }
\label{tab}
\end{table}

\subsection{Ablation analysis}
Since the experimental results in Table \ref{rh36m} and Table \ref{tcmu} have shown the effectiveness of our unified model for both short-term and long-term predictions. Here, we quantitatively analyze other main contributions of the proposed method, including semi-decoupled spatiotemporal modeling, rich multi-granularity modeling.

For the evaluation of semi-decoupled spatiotemporal modeling,
since our model captures motion dynamics in a repeated semi-decoupled manner in the whole process and it is hard to evaluate this via simple modifications,
we compare our results with the related baselines, including the methods that focus on coupled spatiotemporal modeling (i.e. RRNN and ConvSeq) and the method that focus on decoupled spatiotemporal modeling (i.e. LTD). As is shown in Table \ref{rh36m}. Compared with the RRNN and ConvSeq, the errors of our method significantly decrease at all time-steps, showing the effectiveness of our method powerfully. Compared with LTD, our method also achieve the best performance for both short-term and long-term predictions, proving the effectiveness of our method on semi-decoupled spatiotemporal modeling to a great extent.

For the evaluation of rich multi-granularity modeling, we conduct two experiments to show the effectiveness of temporal evolutional directions and aggregation of multiple time scales, respectively: (1) without temporal evolutional directions (w/o TED): we replace the concatenation operations and $k \times k$ Conv layers in Figure \ref{amta} by the element-wise summation. In this case, the network ignores modeling temporal evolutional directions, and therefore we can show its effectiveness. (2) Without aggregation of multi-granularity (w/o AMG): we remove the structure for aggregation of multi-granularity in Figure \ref{amta} to verify its effectiveness. The experimental results are shown in Table \ref{tab}. Compared with the results of ``SDMTL'', the errors of both ``w/o TED'' and ``w/o AMG'' increase at all time-steps, proving the effectiveness of modeling temporal evolutional directions and aggregation of multi-granularity, respectively.

\section{Conclusion}
In this paper, we propose a novel module, BSME, that captures spatiotemporal features from a new perspective in a semi-decoupled manner, and it can be also used to build other spatiotemporal networks. Based on the BSME, we build a novel end-to-end feedforward network, SDMTL, for predicting 3D human motion. In contrast to prior methods,
we capture rich multi-granularity information by summarizing temporal dynamics with adaptive granularities according to the length of the inputs and modeling temporal evolutional directions at each granularity. Extensive experiments on two benchmarks show the effectiveness of semi-decoupled spatiotemporal modeling and rich multi-granularity modeling. In the future, we will focus on exploring the information on multiple spatiotemporal scales, aiming to achieve more accurate predictions at a longer time horizon.

\section{ Acknowledgments}
This work was supported partly by the National Natural Science Foundation of China (Grant No. 61673192), and the Fundamental Research Funds for the Central Universities (Grant No. 2020XD-A04-1, 2019RC27).

\bibliographystyle{aaai21}
\bibliography{SDMTL}

\section{Supplementary Material}

\subsection{Datasets}
{\bf Human3.6M (H3.6M)} \cite{h36m}: H3.6M is the most widely used dataset for 3D human motion prediction, containing 7 actors performing 15 daily activities such as walking and smoking. Each pose has 32 joints. Consistent with the baselines, the sequence of subject 5 (S5) is used for testing, the sequence of subject 11 (S11) is used for validation, and the rest for training.


\noindent{ \bf3D Pose in the Wild dataset (3DPW)} \cite{3dpw}: 3DPW is a new dataset collected in challenging outdoor scenes with a moving camera, including walking in the city, going up-stairs, etc, and it consists of 20 actions performed by 10 subjects. Each pose is represented by 24 joints. We adopt the official training, testing, and validation sets for experiments.

\subsection{Implementation Details}
In experiments, consistent with the baselines, we exclude the global rotation, translation, and constant angles of the 3D coordinate data. The results of ``LTD-long'' in Table 2 and Figure 5 in our manuscript are reproduced using the available code \cite{LTD}.

\subsection{Comparison with  state-of-the-art}
In this section, we provide more results on 3DPW, as is shown in Table \ref{r3dpw}.
Here, the results of LTD in Table \ref{r3dpw} are provided in \cite{LTD}, and their short-term and long-term predictions are obtained by different models with different settings, but the results of our method (i.e. SDMTL-long) for both short-term and long-term predictions are obtained by a unified model.
Compared with the baseline \cite{LTD}, the errors of our method decrease at all time-steps, showing the effectiveness of our method. Our superior performance benefits from two-fold: (1) in contrast to LTD that separately captures the spatial and temporal features, our proposed BSME captures spatiotemporal features in a semi-decoupled manner which is consistent with the cognitive mechanisms of the brain \cite{ecocean,scs}. (2) We capture rich motion dynamics by both summarizing temporal dynamics at multi-granularity and modeling temporal evolutional directions at each granularity.

\begin{table}[h]
\scriptsize
\centering
\begin{tabular}{ccccccc}
\hline
{Milliseconds} & 200 &400 & 600 &800 & 1000 & average\\
 \hline
{LTD} & 35.6& 67.8& 90.6& 106.9& 117.8 & 83.7\\
{SDMTL-long (Ours)} &{\bf 31.7} &{\bf 63.7} &{\bf 88.0}&{\bf 104.6} &{\bf 113.9}& {\bf 80.4}\\
 \hline
\end{tabular}
\caption{Short and long-term predictions on 3DPW. }
\label{r3dpw}
\end{table}

\subsection{Ablation analysis}
\begin{figure}[t]
\centering  
\includegraphics[width=0.55\columnwidth,height=1.35in, trim =2mm 0mm 12mm 8mm, clip=true]{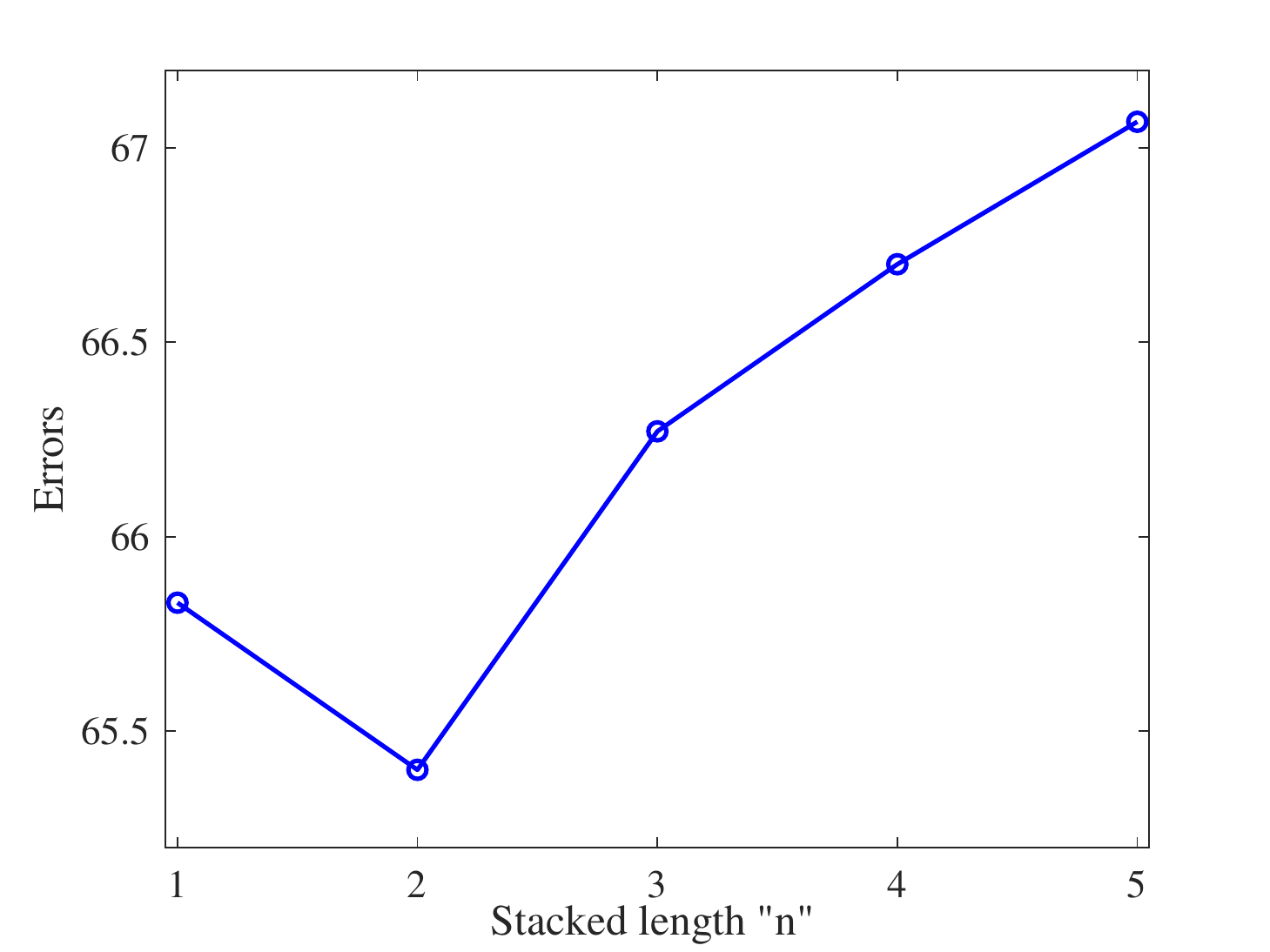} 
\caption{Results for stacking various length (i.e. $n$) of SEs in BSME.}
\label{sl}
\end{figure}

For the evaluation of BSME, we show the effectiveness of the residual connections (RC), stacking multiple SEs in BSME, and the extra interface $X^l_i$ (EI) in Figure 2 of our manuscript, respectively. The results are reported in Table \ref{tab} and Figure \ref{sl}.
(1) Without the RC (i.e. w/o RC), the network performance declines greatly, verifying the effectiveness of capturing multi-grained features of human motion with RC for accurate predictions. Because the RC in BSME can help obtain multi-grained features by gradually fusing the axis-level features from the low layers. (2) Figure \ref{sl} reports the results with various stacked length of SEs, i.e. $n$, ranging from 1 to 5. When $n=2$, we can achieve the best performance. These results can be analyzed from two folds: (a) stacking multiple SEs with proper length can enhance the network to capture the semi-decoupled spatiotemporal features for accurate predictions, i.e. $n=2$ works the best; (b) but stacking too much SEs easily suffers from the problem of over-fitting, leading to the worse performance.
 (3) Without EI, the errors also decrease at all time-steps, showing the effectiveness of fusing spatial features from lower level via the EI.

\begin{table}[h]
\scriptsize
\centering
\begin{tabular}{cccccccc}
\hline
\multirow{2}{*}{Milliseconds}&\multicolumn{7}{c}{Evaluation of BSME} \\
\cline{2-8}
 {} & 80 &160 & 320 &400 &560 & 1000 & average\\
 \hline
 {w/o RC}&10.3 &23.7 &50.8 & 60.7 & 76.9 & 111.4 &55.6 \\
 {w/o EI} & 10.1 & 23.2 & 49.3 & 59.3 & 74.8 & 111.0  & 54.6 \\
 {SDML (Ours)}&{\bf 9.8} & {\bf 22.7} & {\bf 48.0} & {\bf 58.2} & {\bf 74.5}&{\bf 110.7} &{\bf 54.0}\\
 \hline
\hline
 \multirow{2}{*}{Milliseconds}&\multicolumn{7}{c}{Evaluation of loss} \\
 \cline{2-8}
 {} & 80 &160 & 320 &400 &560 & 1000 & average\\
\hline
TW-MPJPE-LW & 11.0 & 24.4 & 50.3 & 60.3 & 76.0 & {\bf 109.8} &55.3\\
MPJPE & 12.1 & 26.5 & 52.5 & 62.0 & 77.6 & 111.7 & 57.1 \\
 {TW-MPJPE (Ours)}&{\bf 9.8} & {\bf 22.7} & {\bf 48.0} & {\bf 58.2} & {\bf 74.5}&{110.7} &{\bf 54.0}\\
\hline
\end{tabular}
\caption{Results of ablative experiments on H3.6M. }
\label{tab}
\end{table}

\begin{figure}[h]
\centering  
\includegraphics[width=0.9\columnwidth,height=1.35in, trim =20mm 0mm 30mm 12mm, clip=true]{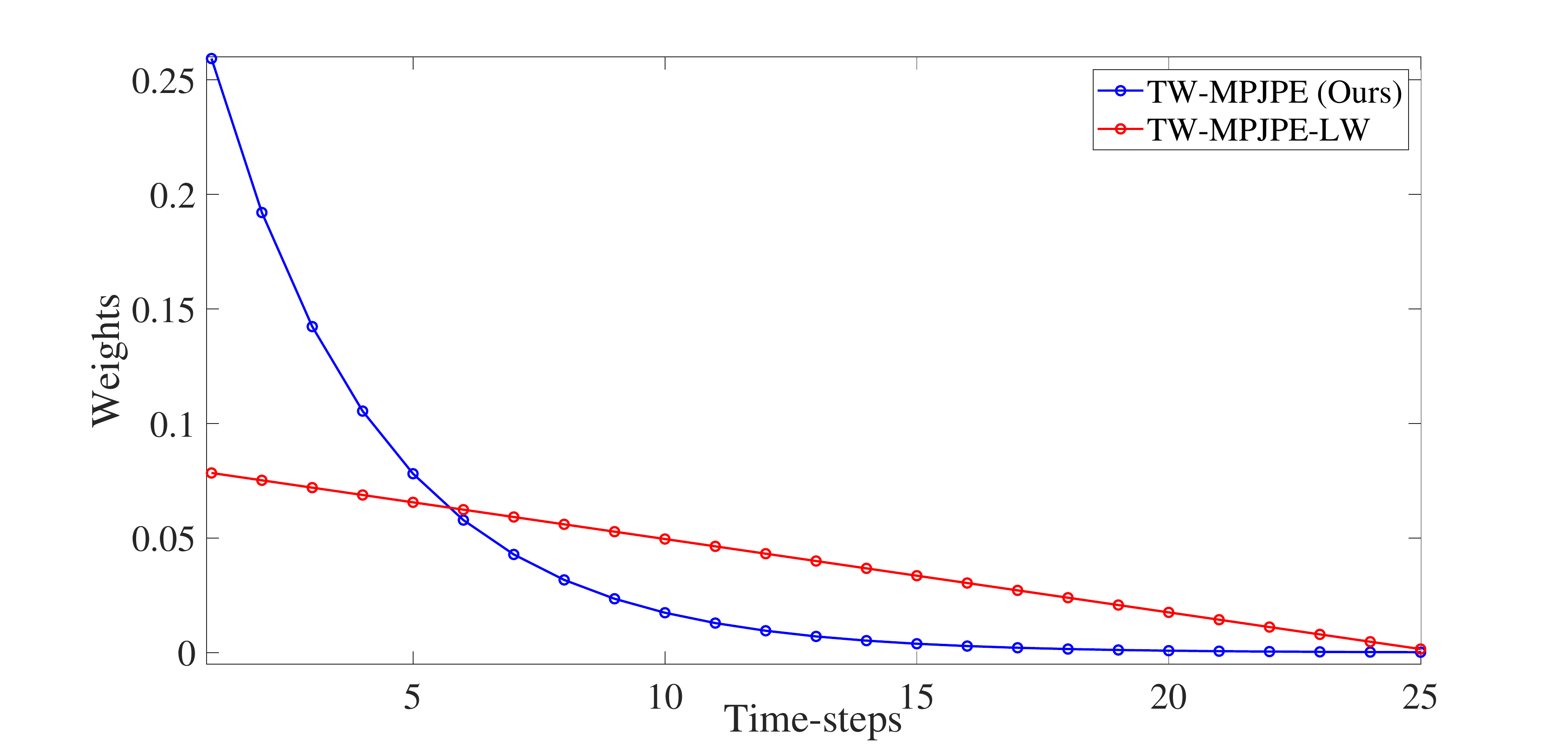} 
\caption{Visualization of attention weights of loss.}
\label{atweights}
\end{figure}

For the evaluation of loss, we conduct two experiments: (1) TW-MPJPE with linearly decreasing weights (TW-MPJPE-LW): in this case, the weight ${w_i}$ is linearly decreasing with the time-steps of the predictions. (2) Remove the decreasing weights (MPJPE): in this experiment, we did not use attention weights for guiding predictions. Instead, we use the Mean Per Joints Position Error (MPJPE) loss which pays equal attention for all time-steps.
The attention weights of the loss ``TW-MPJPE-LW'' and ``TW-MPJPE (Ours)'' are shown in Figure \ref{atweights}.
Using the linearly decreasing weights, the results of ``TW-MPJPE-LW'' decline on average, showing the effectiveness of our loss using nonlinearly decreasing weights which is consistent with the difficulty of predictive tasks. By contrast, without the decreasing weights of the loss, the errors increase at all time-steps, verifying the effectiveness of the attention weights for guiding accurate predictions.

\end{document}